\documentclass[conference]{IEEEtran} 
\IEEEoverridecommandlockouts 

\usepackage{cite}
\usepackage{amsmath,amssymb,amsfonts}
\usepackage{algorithmic}
\usepackage{graphicx}
\usepackage{textcomp}
\usepackage{xcolor}
\usepackage{url}
\usepackage{amsmath}
\usepackage{amssymb}
\usepackage{bbm}
\usepackage{dsfont}
\usepackage[linesnumbered,ruled,vlined]{algorithm2e}
\SetKwInput{KwInput}{Input}     
\SetKwInput{KwOutput}{Output}     
\usepackage{mathtools}
\usepackage{multirow}

\def\BibTeX{{\rm B\kern-.05em{\sc i\kern-.025em b}\kern-.08em
    T\kern-.1667em\lower.7ex\hbox{E}\kern-.125emX}}
    
\title{Importance Reweighting for Biquality Learning}

\author{
\IEEEauthorblockN{Pierre Nodet}
\IEEEauthorblockA{
\textit{Orange Labs} \\
\textit{AgroParisTech, INRAe}\\
46 av. de la République\\
Châtillon, France}
\and
\IEEEauthorblockN{Vincent Lemaire}
\IEEEauthorblockA{\textit{Orange Labs} \\
2 av. P. Marzin\\
Lannion, France}
\and
\IEEEauthorblockN{Alexis Bondu}
\IEEEauthorblockA{\textit{Orange Labs} \\
46 av. de la République\\
Châtillon, France}
\and
\IEEEauthorblockN{Antoine Cornuéjols}
\IEEEauthorblockA{\textit{UMR MIA-Paris}\\ 
\textit{AgroParisTech, INRAe}\\
\textit{Universit\'e Paris-Saclay}\\
16 r. Claude Bernard\\
Paris, France}
}

\begin{document}

\maketitle

\begin{abstract}

The field of Weakly Supervised Learning (WSL) has recently seen a surge of popularity, with numerous papers addressing different types of
``supervision deficiencies'', namely: poor quality, non adaptability, and insufficient quantity of labels. Regarding quality,
label noise can be of different types, including completely-at-random, at-random or even not-at-random.
All these kinds of label noise are addressed separately in the literature, leading to highly specialized approaches.
This paper proposes an original, encompassing, view of Weakly Supervised Learning, which results in the
 design of generic approaches capable of dealing with any kind of label noise. For this purpose, an alternative setting called ``Biquality data'' is used.
It assumes that a small trusted dataset of correctly labeled examples is available, in addition to an untrusted dataset of noisy examples.
In this paper, we propose a new reweigthing scheme capable of identifying noncorrupted examples in the untrusted dataset.
This allows one to learn classifiers using both datasets. Extensive experiments that simulate several types of label noise and that vary the quality
and quantity of untrusted examples, demonstrate that the proposed approach outperforms baselines and state-of-the-art approaches.
\end{abstract}


\section{Introduction}

The supervised classification problem aims to learn a classifier from a set of labeled training examples in order to predict the class of new examples. 
In practice, conventional classification techniques may fail because of the imperfections of real-world datasets. 
Accordingly, the field of \textit{Weakly Supervised Learning} (WSL) has recently seen a surge of popularity, with numerous papers addressing different types of \textit{``supervision deficiencies''} \cite{zhou:2017}, namely:

{\bf Insufficient quantity:} when many training examples are available, but only a small portion is labeled, e.g. due to the cost of labelling. For instance, this occurs in the field of cyber security where human forensics is needed to label attacks. Usually, this issue is addressed by semi-supervised learning (SSL) \cite{chapelle2009semi} or active learning (AL) \cite{settles2009active}.

{\bf Poor  quality labels}: in this case, all {training examples} are labeled but the labels may be corrupted.
This may happen when the labeling task is outsourced to crowd labeling. The Robust Learning to Label Noise (RLL) approaches address this
problem \cite{frenay2013classification}, with three identified types of label noise: i) the \textit{completely at random}
noise which correspond to a uniform probability of label change~; ii) the \textit{at-random} label noise when the probability
of label change depends upon each class, with uniform label changes within each class~; iii) the \textit{not-at-random} label noise
 when the probability of label change varies over the input space of the classifier. 
This last type of label noise is recognized as the most difficult to deal with \cite{cheng2017learning,menon2016learning}.

{\bf Inappropriate labels}:
for instance, in Multi Instance Learning (MIL) \cite{Carbonneau_2018} the labels are assigned to bags of examples, with positive label indicating that at least one example of the bag is positive.
Some scenarios in  Transfer Learning (TL) \cite{weiss2016survey} imply that only the labels in the source domain are provided while the target domain labels are not. Often, these non-adapted labels are associated with slightly different learning tasks \textit{(e.g. more precise and numerous classes are dividing the original categories)}.
Alternatively, non-adapted labels may characterize a differing statistical individual \cite{Conte2010} \textit{(e.g. a subpart of an image instead of the entire image)}. 

All these types of supervision deficiencies are addressed separately in the literature, leading to highly specialized approaches. 
In practice, it is very difficult to identify the type(s) of deficiencies with which a real dataset is associated. 
For this reason, we argue that it would be very useful to find a unified framework for Weakly Supervised Learning, in order to design generic approaches capable of dealing with any type of supervision deficiency. 

In Section \ref{biQ-data} of this paper, we present \textit{``biquality data''}, an alternative WSL setting allowing a unified view of weakly supervised learning.
A generic learning framework using the two training sets of biquality data \textit{(the one trusted and the other one untrusted)}
 is suggested in Section \ref{biQ-learn}.  
We identify three possible ways of implementing this framework and consider one of them. 
This article proposes a new approach using example reweighting in Section \ref{new-approach}. 
The effectiveness of this new approach in dealing with different types of supervision deficiencies, without a priori knowledge about them, is demonstrated through experiments with real datasets in Sections \ref{description_expe} and \ref{sec_results_analysis}. 
Finally, perspectives and future works are discussed in Section \ref{sec_conclusion}. 

\section{Biquality Data}
\label{biQ-data}

This section presents an alternative setting called \textit{``Biquality Data''} which covers a large range of supervision deficiencies and allows for unifying the WSL approaches. The interested reader may find a more detailed introduction on WSL and its links with Biquality Learning in  \cite{nodet2021weakly}.

Learning using biquality data has recently been put forward in \cite{Charikar2016,Hendrycks2018,Hataya2019UnifyingSA} and consists in learning a classifier from two distinct training sets, one trusted and the other untrusted. The initial motivation was to unify semi-supervised and robust learning through a combination  of the two.  We show in this paper that this scenario is not limited to this unification and that it can cover a larger range of supervision deficiencies as demonstrated with the algorithms we propose and the obtained results. 

The trusted dataset $D_T$ consists of pairs of labeled examples ($x_i, y_i$) where all labels $y_i \in \mathcal{Y}$ are supposed to be correct according to the true underlying conditional distribution $\mathbb{P}_T(Y|X)$. In the untrusted dataset $D_U$, examples $x_i$ may be associated with incorrect labels. We note $\mathbb{P}_U(Y|X)$ the corresponding conditional distribution.

At this stage, no assumption is made about the nature of the supervision deficiencies which could be of any type including label noise,
missing labels, concept drift, non-adapted labels... and more generally a mixture of these supervision deficiencies.

The difficulty of a learning task performed on biquality data can be characterised by two quantities.
First, the ratio of trusted data over the whole data set, denoted by $p$: 
\begin{equation}
p = \frac {|D_T|}{|D_T|+|D_U|}
\end{equation}

Second, a measure of the quality, denoted by $q$, which evaluates the usefulness of the untrusted data $D_U$ to learn the trusted concept $\mathbb{P}_T(Y|X)$, where $q\in [0,1]$ and 1 indicates high quality. For example in \cite{Hataya2019UnifyingSA} $q$ is defined using a ratio of Kullback-Leibler divergence between $\mathbb{P}_T(Y|X)$ and $\mathbb{P}_U(Y|X)$.

\begin{figure}[!ht]
\centering
\includegraphics[width=\linewidth]{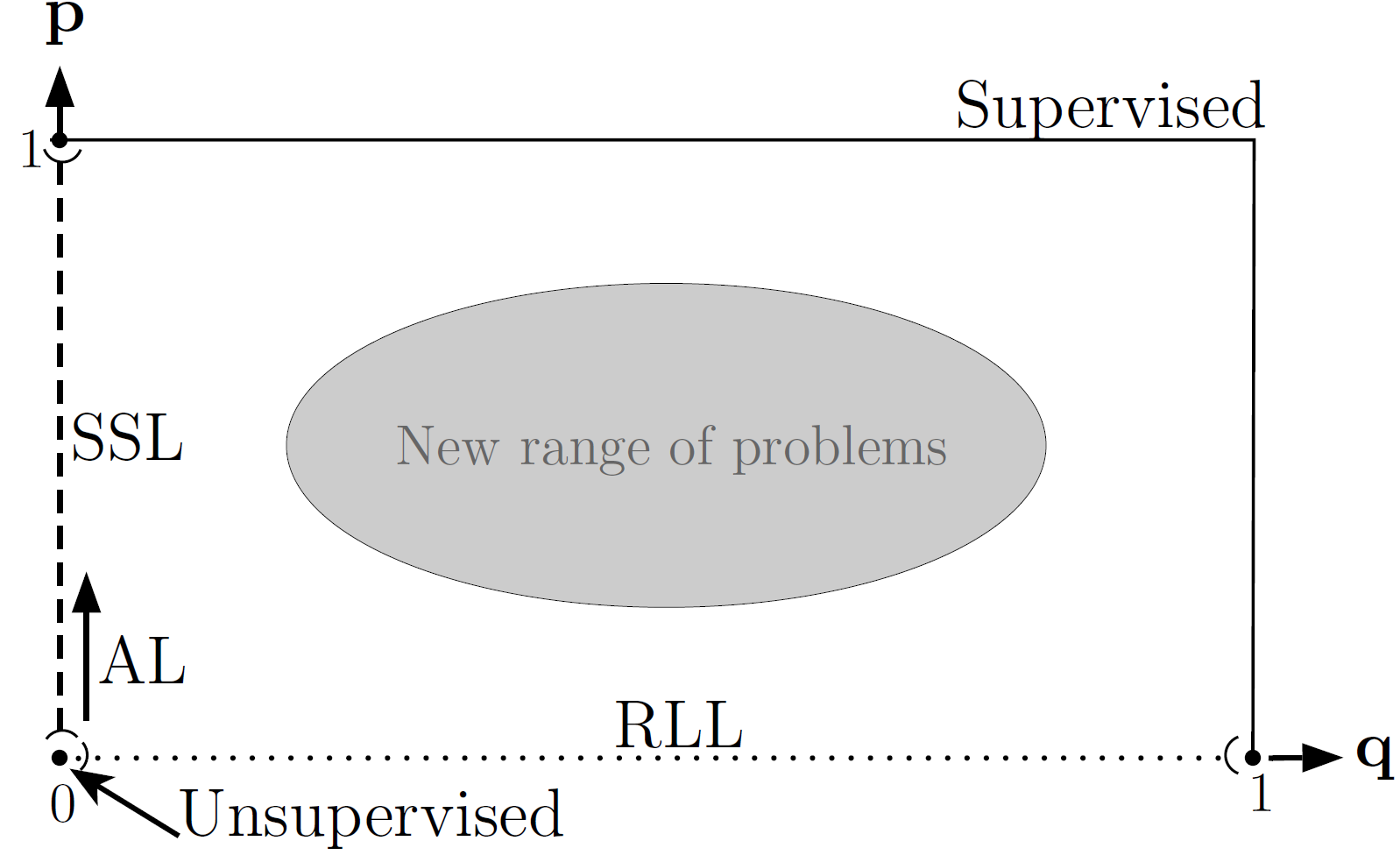}
\caption{The different learning tasks covered by the biquality setting, represented on a 2D representation.}
\label{2D}
\end{figure}

The biquality setting covers a wide range of learning tasks by varying the quantities $q$ and $p$  (as represented in Figure \ref{2D}):  

\begin{itemize}
  \item When ($p=1$ OR $q=1$)\footnote{~$p=1 \implies D_U=\emptyset \implies q=1$}  all examples can be trusted. Thus, this setting corresponds to a standard {\bf supervised learning} (SL) task.
  
  \item When ($p=0$ AND $q=0$), there is no trusted {examples} and the untrusted labels are not informative. We are left with only the inputs $\{x_i\}_{1 \leq i \leq m}$ as in {\bf unsupervised learning (UL)}.
  
  \item On {the vertical axis defined by} $q=0$, except for the two points $(p,q)=(0,0)$ and $(p,q)=(1,0)$, the untrusted labels are not informative, and trusted examples are available. The learning task becomes {\bf semi-supervised learning (SSL)} with the untrusted examples as unlabeled and the trusted as labeled.
  
  \item An upward move on the vertical axis, from a point $(p,q)=(\epsilon,0)$ characterized by a low proportion of labeled examples $p=\epsilon$, to a point $(p',0)$, with $p' > p$, corresponds to {\bf Active Learning}, when an oracle provides new labels asked by a given strategy. The same upward move can also be realized in {\bf Self-training} and {\bf Co-training} \cite{ZHAO2017}, where unlabeled training examples are labeled using the predictions of the current classifier(s).
  
  \item On the horizontal axis {defined by} $p=0$, {except for the points} $(p,q)=(0,0)$ and $(p,q)=(0,1)$, only untrusted examples are provided, which corresponds to the range of learning tasks typically addressed by {\bf Robust Learning to Label noise (RLL)} approaches.\\ 
  \end{itemize}

Only the edges of Figure \ref{2D} have been envisioned in previous works {--  i.e.} the points mentioned above {--} and a whole new range of problems are addressed in this paper.
Moreover, biquality learning may be used to tackle tasks belonging to WSL, for instance:

\begin{itemize}
  \item Positive Unlabeled Learning (PUL) \cite{Bekker_2020}
  where only positive (trusted) and unlabeled instances are available, the later which can be considered as untrusted. 

  \item Self Training and Cotraining \cite{ZHAO2017} could be addressed at the end of the self labeling process: the initial training set
  is then the trusted dataset, and all self-labeled examples can be considered as the untrusted ones. 
  
  \item Concept drift \cite{Gama2014}: when a concept drift occurs, all the examples used before a detected drift may be considered as the untrusted examples, while the examples available after it are viewed as the trusted ones, assuming a perfect labeling process. 
  
  \item Self Supervised Learning system as illustrated by Snorkel \cite{ratner2020snorkel} or Snuba \cite{snuba}: the small initial training set can the trusted, whereas all examples automatically labeled using the labeling functions may be considered as untrusted.
\end{itemize}

As can be seen from the above list, the Biquality framework is quite general and its investigation seems a promising avenue to unify different aspects of the  Weakly Supervised Learning. A main contribution of this paper is to suggest one generic framework for achieving biquality learning and thus covering many facets of WSL. This is presented in the next section.
This framework will be then applied in the experiments part of this paper to the problem of label noise.

\section{Biquality Learning}
\label{biQ-learn}

Learning the true concept\footnote{For reasons of space, we denote $\mathbb{P}_T(Y|X)$  by $T$ and $\mathbb{P}_U(Y|X)$ by $U$.} $\mathbb{P}_T(Y|X)$ on $D = D_T \cup D_U$ means minimizing the risk $R$ on $D$ with a loss $L$ for a probabilistic classifier $f$:
\begin{equation}
\small{
\begin{aligned}
R_{D,L}(f) &= \mathbb{E}_{D,(X,Y)\sim T}[L(f(X),Y)]\\
&= \mathbb{P}(X \in D_T)\mathbb{E}_{D_T,(X,Y)\sim T}[L(f(X),Y)]\\
&+ \mathbb{P}(X \in D_U)\mathbb{E}_{D_U,(X,Y)\sim T}[L(f(X),Y)]\\
\end{aligned}
}
\label{risk-all-data}
\end{equation}
where $L(\cdot, \cdot)$ is a loss function, from $\mathbb{R}^{|\mathcal{Y}|}\times\mathcal{Y}$ to $\mathbb{R}$ since $f(X)$ is a vector of probability over the classes. 
Since the true concept $\mathbb{P}_T(Y|X)$ cannot be learned from $D_U$, the last line of Equation \ref{risk-all-data} is not tractable as it stands.
That is why we propose a {\bf generic formalization} based on a mapping function $g$ that enables us to learn the true concept from the modified untrusted examples of $D_U$. Equation \ref{risk-all-data} becomes:
\begin{equation}
\small{
\begin{aligned}
R_{D,L}({f}) &= \mathbb{P}(X \in D_T)\mathbb{E}_{D_T,(X,Y)\sim T}[L(f(X),Y)]\\
&+ \lambda \, \mathbb{P}(X \in D_U)\mathbb{E}_{D_U,(X,Y)\sim U}{[g(L(f(X),Y))]}\\
\end{aligned}
}
\label{risk-with-g}
\end{equation}

In Equation \ref{risk-with-g}, the parameter $\lambda \in [0,1]$ reflects the quality of the untrusted examples of $D_U$ modified by the function $g$. 
This time, the last line is tractable since it consists of a risk expectancy estimated over the training examples of $D_U$ which follows the untrusted concept $\mathbb{P}_U(Y|X)$, modified by the function $g$. 

Accordingly, the estimation of the expected risk requires to learn three items: $g$, $\lambda$ and then $f$. 
To learn $g$, a mapping function between the two datasets, both $D_T$ and $D_U$ are used.
Then, either $\lambda$ is considered as a hyper parameter to be learned using $D_T$ or $\lambda$ is provided by an appropriate quality measure and is considered as an input of the learning algorithm.
Finally, $f$ is learned by minimizing the risk $R$ on $D$ using the mapping $g$. 

In this formalization, the mapping function $g$ plays a central role. 
Not exhaustively, we identify three different ways of designing the mapping function. For each of these, a different function $g'$ enters the definition of function $g$:
\begin{itemize}
  \item The first option consists in {\bf correcting the label} for each untrusted examples of $D_U$. The mapping function thus takes the form $g(L(f(X),Y))= L(f(X),g'(Y,X))$, with $g'(Y,X)$ denoting the new corrected labels and $f(X)$ the predictions of the classifier.
  \item In the second option, the untrusted labels are used unchanged. The untrusted examples $X$ are {\bf moved} in the input space where the untrusted labels becomes correct with respect to the true underlying concept. The mapping function becomes $g(L(f(X),Y))= L(f(g'(X)),Y)$, where $g'(X)$ is the ``moved'' input vector of the modified untrusted examples.
  \item In the last option, $g'$ {\bf weights} the contribution of the untrusted examples in the risk estimate. Accordingly, we have $g(L(f(X),Y))= g'(Y,X)L(f(X),Y)$. In this case, the parameter $\lambda$ may disappear from Equation \ref{risk-with-g} since it can be considered as included in the function $g'$.
\end{itemize}

Section \ref{new-approach} considers in-depth the last option and proposes a new approach where $g'$ acts as an Importance Reweighting for Biquality Learning.

\section{A new Importance Reweighting approach for Biquality Learning}
\label{new-approach}

To estimate the mapping function $g'$, we suggest to adapt the importance reweigthing trick from the covariate shift literature \cite{Liu_2016} to biquality learning.
This trick relies on reweighting untrusted samples by using the Radon-Nikodym derivative (RND) \cite{Nikodym1930} of $\mathbb{P}_T(X,Y)$ in respect to $\mathbb{P}_U(X,Y)$ which is $\frac{\text{d}\mathbb{P}_T(X,Y)}{\text{d}\mathbb{P}_U(X,Y)}$.
Contrary to the ``covariate shift'' setting, the biquality setting handles the same distribution $\mathbb{P}(X)$ in the trusted and untrusted datasets. However, the two underlying concepts $\mathbb{P}_T(Y|X)$ and $\mathbb{P}_U(Y|X)$ are possibly different due to a supervision deficiency.
By using these assumptions and the Bayes Formula, we can further simplifying the reweighing function to the RND of $\mathbb{P}_T(Y|X)$ in respect to $\mathbb{P}_U(Y|X)$, $\frac{\text{d}\mathbb{P}_T(Y|X)}{\text{d}\mathbb{P}_U(Y|X)}$.

\begin{equation}
\begin{aligned}
&R_{(X,Y)\sim T,L}(f) = \mathbb{E}_{(X,Y)\sim T}[L(f(X),Y)]\\
&\qquad = \int L(f(X),Y)\,\text{d}\mathbb{P}_T(X,Y)\\
&\qquad = \int \frac{\text{d}\mathbb{P}_T(X,Y)}{\text{d}\mathbb{P}_U(X,Y)}L(f(X),Y)\,\text{d}\mathbb{P}_U(X,Y)\\
&\qquad = \mathbb{E}_{(X,Y)\sim U}[\frac{\mathbb{P}_T(X,Y)}{\mathbb{P}_U(X,Y)}L(f(X),Y)]\\
&\qquad = \mathbb{E}_{(X,Y)\sim U}[\frac{\mathbb{P}_T(Y|X)\mathbb{P}(X)}{\mathbb{P}_U(Y|X)\mathbb{P}(X)}L(f(X),Y)]\\
&\qquad = \mathbb{E}_{(X,Y)\sim U}[\frac{\mathbb{P}_T(Y|X)}{\mathbb{P}_U(Y|X)}L(f(X),Y)]\\
&\qquad = \mathbb{E}_{(X,Y)\sim U}[\beta L(f(X),Y)]\\
&\qquad = R_{(X,Y)\sim U,\beta L}(f)\\
\end{aligned}
\label{risk-loss}
\end{equation}

Equation \ref{risk-loss} shows that $\beta = \frac{\mathbb{P}_T(Y|X)}{\mathbb{P}_U(Y|X)}$ is an estimation of the mapping function $g'$, thanks to Section \ref{biQ-learn} estimating $\beta$ is the last step before an actual Biquality Learning algorithm.

\begin{algorithm}[!h]
\renewcommand{\thealgocf}{}
\DontPrintSemicolon
  \KwInput{Trusted Dataset $D_T$, Untrusted Dataset $D_U$, Probabilistic Classifier Familiy $\mathcal{F}$}
  Learn $f_U \in \mathcal{F}$ on $D_U$\;
  Learn $f_T \in \mathcal{F}$ on $D_T$\;
  \For{$(x_i,y_i) \in D_U$, where $y_i \in [\![1,K]\!]$ }
    { 
    	$\hat{\beta}(x_i,y_i) = 
    	 \left < \frac{f_T(x_i)}{f_U(x_i)}  \right >_{y_i}$
    }
  \For{$(x_i,y_i) \in D_T$}  
    { 
    	$\hat{\beta}(x_i,y_i) = 1$
    }
  Learn $f \in \mathcal{F}$ on $D_T \cup D_U$ with weights $\hat{\beta}$ \\
  \KwOutput{$f$}
\caption{Importance Reweighting for Biquality Learning (IRBL)}
\label{algo}
\end{algorithm}

The proposed algorithm, Importance Reweighting for Biquality Learning (IRBL), aims at estimating $\beta$ from $D_T$ and $D_U$ whatever the unknown supervision deficiency. It consists of two successive steps. First a probabilistic classifier $f_T$ is learned from the trusted dataset $D_T$ and another probabilistic classifier $f_U$ is learned from the untrusted dataset $D_U$. Thanks to their probabilistic nature each of them estimates $\mathbb{P}_T(Y|X)$ and $\mathbb{P}_U(Y|X)$ by a probability distribution over the set of the $K$ classes. Thus we can estimate the weight $\beta$ of an untrusted sample $(x_i,y_i)$ by dividing the prediction of $f_T(x_i)$ by $f_U(x_i)$ for the $y_i$ class \textit{(see line 4)}. The weight $\beta$ for all trusted samples will be fixed to 1 \textit{(see line 6)}. Then a final classifier is learned from both datasets $D_T$ and $D_U$ with examples reweighted by $\hat{\beta}$.

Our algorithm is theoretically grounded, since it is asymptotically equivalent to minimizing the risk on the true concept using the entire data set (see proof in Equation \ref{proof}).

\begin{equation}
\begin{aligned}
\hat{R}_{D,\hat{\beta}L}(f) &= \frac{1}{|D|}\sum_{(x_i,y_i) \in D}\big(\mathds{1}_{(x_i,y_i) \in D_T}L(f(x_i),y_i)\\
&+ \mathds{1}_{(x_i,y_i) \in D_U}\hat{\beta}(x_i,y_i)L(f(x_i),y_i)\big)\\
&= \frac{1}{|D_T|+|D_U|}\sum_{(x_i,y_i) \in D_T}L(f(x_i),y_i)\\
&+ \frac{1}{|D_T|+|D_U|}\sum_{(x_i,y_i) \in D_U}L(f(x_i),y_i)\hat{\beta}(x_i,y_i)\\
&= \frac{p}{|D_T|}\sum_{(x_i,y_i) \in D_T}L(f(x_i),y_i)\\
&+ \frac{1-p}{|D_U|}\sum_{(x_i,y_i) \in D_U}L(f(x_i),y_i)\hat{\beta}(x_i,y_i)\\
&= p\hat{R}_{D_T,L}(f) + (1-p)\hat{R}_{D_U,\hat{\beta}L}(f)\\
&\approx p\hat{R}_{D_T,L}(f) + (1-p)\hat{R}_{D_T,L}(f)\\
&\approx \hat{R}_{D_T,L}(f)
\end{aligned}
\label{proof}
\end{equation}

Proof in Equation \ref{proof} is an asymptotic result: in practice our algorithm relies on the quality of the estimation of $\mathbb{P}_T(Y|X)$ and $\mathbb{P}_U(Y|X)$ in order to be efficient. In the biquality setting they both could be hard to estimate because of the small size of $D_T$ and the poor quality of $D_U$. 

\section{Experiments}
\label{description_expe}

The aim of the experiments is to answer the following questions: i) is our algorithm properly designed and does it perform better than baselines approaches? ii) is our algorithm competitive with state-of-the-art approaches?

First, Section \ref{SupervisionDeficiencies} presents the supervision deficiencies which are simulated in our experiments. They correspond to two different kinds of weak supervision, namely, 
Noisy label Completely at Random (i.e. not $X$ dependent) and Noisy label Not at Random (i.e. $X$ dependent). From the Frenay's taxonomy \cite{frenay2013classification} the former is the easiest to deal with and the later is often considered as  difficult to manage. Then, Section \ref{competitors} consists of three parts: a presentation of the baseline competitors, a brief report on the state-of-the-art competitors, and a description of the set of classifiers used. Finally, Section \ref{data_and_clf} describes the datasets used in the experiments, and the chosen criterion to evaluate the learned classifiers.
For full reproducibility, source code, datasets and results are available at : \url{https://github.com/pierrenodet/irbl}.

\subsection{Simulated supervision deficiencies}
\label{SupervisionDeficiencies}

The datasets listed in Section \ref{data_and_clf} consist of a collection of training examples that are assumed to be correctly labeled, denoted by $D_{total}$. 
In  order  to  obtain  a  trusted  dataset $D_T$ and  an  untrusted one $D_U$, each dataset is split in two parts using a stratified random draw, where $p$ is the percentage for the trusted part.
The trusted datasets are left untouched, whereas corrupted labels are simulated in the untrusted datasets by using two different techniques:

\paragraph{Noisy Completely At Random (NCAR):} Corrupted untrusted examples are uniformly drawn from $D_U$ with a probability $r$, and are assigned a random label that is also uniformly drawn from $\mathcal{Y}$. 

In the particular case of binary classification problems, the conditional distribution of the untrusted labels is defined by Equation \ref{mcar1}.  
\begin{equation}
\forall y \in \mathcal{Y}, \mathbb{P}_U(Y=y|X) = \frac{r}{2} + (1-r)\mathbb{P}_T(Y=y|X)
\label{mcar1}
\end{equation}
Here, $r$ controls the overall number of random labels and thus is our proxy for the quality: $q=1-r$.

\paragraph{Noisy Not At Random (NNAR):}

Corrupted untrusted examples are drawn from $D_U$ with a probability $r(x)$ that depends on the instance value.
To generate a instance dependent label noise, we design a noise that depends on the decision boundary of a classifier $f_{total}$
learned from $D_{total}$. The probability of random label $r(x)$ should be high when an instance $x$ is close to the decision boundary,  and low when it is far. In our experiments, the probability outputs of $f_{total}$ are used to model our label noise as follows:
\begin{equation}
\label{rnar-noise} \forall x \in \mathcal{X}, r(x) =  1 - \theta|1 - 2f_{total}(x)|^{\frac{1}{\theta}}
\end{equation}
where $\theta \in [0;1]$ is a constant that controls the overall number of random labels and thus is our proxy for the quality: $q = \theta$. The parameter $\theta$ influences both the slope (factor) and the curvature (power) of $r(x)$ to modify the area under curve of $r(x)$: $\mathbb{E}[r(x)]$.

For binary classification problems, the conditional distribution of the untrusted labels is defined by Equation \ref{mcar2}.
\begin{equation}
\begin{aligned}
&\forall x \in \mathcal{X},\forall y \in \mathcal{Y},\\
&\mathbb{P}_U(Y\!=\!y|X=x)=\!\frac{r(x)}{2}\!+\!(1\!-r(x))\mathbb{P}_T(Y\!=\!y|X\!=\!x)\\
\end{aligned}
\label{mcar2}
\end{equation}

\subsection{Competitors}
\label{competitors}

\paragraph{Baseline competitors} 
The first part of our experiments consists of a sanity check which compares the performance of the proposed algorithm to the following baselines:

\begin{itemize}
  \item \textit{Trusted}: The final classifier $f$ obtained with our algorithm should be better than a classifier $f_T$ that learned only from the trusted dataset, insofar as untrusted data bring useful information about the trusted concept. At least, $f$ should not be worse than using only trusted data.
  
  \item \textit{Mixed}: The final classifier $f$ should be better than a classifier $f_{mixed}$ learned from both trusted and untrusted dataset, without correction. A biquality learning algorithm should leverage the information provided by having two distinct datasets.
  
  \item \textit{Untrusted}: The final classifier should be better than a classifier $f_U$ that learns only from the untrusted dataset if there are trusted labels. Using trusted data should improve the classifier final performances.\\
\end{itemize}

\paragraph{State-of-the-art-competitors} 
The second part of our experiments compares our algorithm with two state-of-the-art methods: (i) a method from the Robust Learning
to Label noise (RLL) \cite{NIPS2015_5941,charoenphakdee_symmetric_2019} family and (ii)  the GLC approach \cite{Hendrycks2018}. 

\begin{itemize}
\item \textit{RLL}: In recent literature a new emphasis is put on the research of new loss functions that are conducive to better risk minimization
in presence of noisy labels. For example, \cite{NIPS2015_5941,charoenphakdee_symmetric_2019} show theoretically and experimentally that when the loss function satisfies a symmetry condition, described below, this contributes to the robustness of the classifier. Accordingly, in this paper we train a classifier with a symmetric loss function as a competitor. This first competitor is expected to have good results on completely-at-random label noise described in Section \ref{SupervisionDeficiencies}.
A loss function $L_s$ is said \textit{symmetrical}  if $\sum_{y \in \{-1;1\}} L_s(f(x), y)=c$, where $c$ is a constant and $f(x)$ is the score on the class $y$. This loss function is used on $D_T \cup D_U$.\\

\item \textit{GLC}:  To the best of our knowledge, GLC \cite{Hendrycks2018} is among the best performing algorithm that can learn from bi-quality data. It has been successfully compared to many competing approaches. Like ours, it is a two steps approach which is simple and easy to implement. 

In a first step, a model $f_U$ is learned from the untrusted dataset $D_U$. Then it is used to estimate a transition matrix $C$ of $\mathbb{P}_{U|T}(Y)$ by making probabilistic predictions with $f_U$ on the trusted dataset $D_T$ and comparing it to the trusted labels. 

In a second step, this matrix is used to correct the labels from the untrusted dataset $D_U$ when learning the final model $f$. Indeed $f$ is learned with $L$ on $D_T$ and with $L(C^{\top} f(X),Y)$ on $D_U$.\\
\end{itemize}

\paragraph{Classifiers} 
First of all, the choice of classifiers was guided by the idea of comparing algorithms for biquality learning and not searching for the best classifiers.
This choice was also guided by the nature of the datasets used in the experiments (see section \ref{data_and_clf}).
Secondly our algorithm, as well as GLC, implies two learning phases. For both reasons and for simplicity, we decided to use Logistic Regressions (LR) for each phase. LR is known to be limited, in the sense of the Vapnik-Chervonenkis dimension \cite{vapnik1995nature} since it can only learn linear separations of the input space $\mathcal{X}$, which could underfit the conditional probabilities $\mathbb{P}(Y|X)$ on $D_T$ and $D_U$ and lead to bad $\beta$ estimations.  But this impediment, if met, will affect equally all the compared algorithms.
LR is also used for the RLL classifier using the Unhinged symmetric loss function.

To obtain reliable estimations of conditional probabilities $\mathbb{P}(Y|X)$, the outputs of all classifiers have been calibrated thanks to Isotonic Regression with the default parameters provided by scikit-learn \cite{scikit-learn}.

Logistic Regression is always be used and learned thanks to SGD with a learning rate of $0.005$, a weight decay of $10^{-6}$ during $20$ epochs and a batch size of $24$ with Pytorch \cite{NEURIPS2019_9015}.

\subsection{Datasets}
\label{data_and_clf}

In industrial applications familiar to us, such as fraud detection, Customer Relationship Management (CRM) and churn prediction, we are mostly faced with  binary classification problems. The available data is of average size in terms of the number of explanatory variables and involves mixed variables (numerical and categorical).

For this reason 
we limited in this paper the experiments to binary classification tasks even if our algorithm can address multi-class problems.
The chosen tabular datasets, used for the experiments, have similar characteristics than those of our real applications.

\begin{table}[!h]
\caption{Binary classification datasets used for the evaluation. Columns: number of examples ($|D|$), number of features ($|\mathcal{X}|$), and ratio of examples from the minority class (min).}
\centering
\fontsize{7}{8}\selectfont
\begin{tabular}{|l|l|l|l||l|l|l|l|}
\hline
name     & $|D|$ & $|\mathcal{X}|$ & min & name     & $|D|$ & $|\mathcal{X}|$ & min \\ \hline\hline
4class & 862 & 2 & 36 & ibnsina & 20,722 & 92 & 38\\ \hline
ad   & 3,278 & 1558 & 14 & zebra & 61,488 & 154 & 4.6\\ \hline
adult   & 48,842 & 14 & 23 & musk & 6,598 & 169 & 15\\ \hline
aus & 690 & 14 & 44 & phishing & 11055 & 30 & 44\\ \hline
banknote & 1372 & 4 & 44 & spam & 4,601 & 57 & 39\\ \hline
breast & 683 & 9 & 35 & ijcnn1 & 141,691 & 22 & 9\\ \hline
eeg & 1498 & 13 & 45 & svmg3 & 1284 & 4 & 26\\ \hline
diabetes & 768 & 8 & 35 & svmg1 & 7,089 & 22 & 43\\ \hline
german & 1000 & 20 & 30 & sylva & 145,252 & 108 & 6.5\\ \hline
hiva & 42,678 & 1617 & 3.5 & web & 49,749 & 300 & 3\\ \hline
\end{tabular}
\end{table}

They come from different sources:  UCI \cite{Dua2019}, libsvm\footnote{\url{https://www.csie.ntu.edu.tw/~cjlin/libsvmtools/datasets/}} and active learning challenge \cite{guyon2010dataset}. A part of these datasets comes from past challenges on active learning where high performances with a low number of labeled examples has proved difficult to obtain.
For each dataset, 80~\% of samples were used for training and 20\% were used for the test. With this choice of datasets, a large range of the class ratio is covered: Australian is almost balanced while Web is really unbalanced. Also, the size varies significantly in number of rows or columns,  with corresponding impact on the difficulty of the learning tasks.

\section{Results}
\label{sec_results_analysis}

The empirical performance of our approach, is evaluated in two steps. 
First, we investigate the efficiency of the reweighting scheme and its influence on the learning procedure of the final classifier. 
Second, our approach is benchmarked against competitors to evaluate its efficiency in real tasks.

\subsection{Behavior of the IRBL method}

In order to illustrate the proposed reweighing scheme, we picked a dataset, here the ``ad'' dataset used with a ratio of trusted data $p=0.25$, and examined the histogram of the weights assigned to each untrusted example either corrupted or not. The case of Random  Label  Completely  at  Random is chosen and the hardest case where all labels are at random $q=0$ is considered.

\begin{figure}[!h]
\centering
\includegraphics[width=\linewidth]{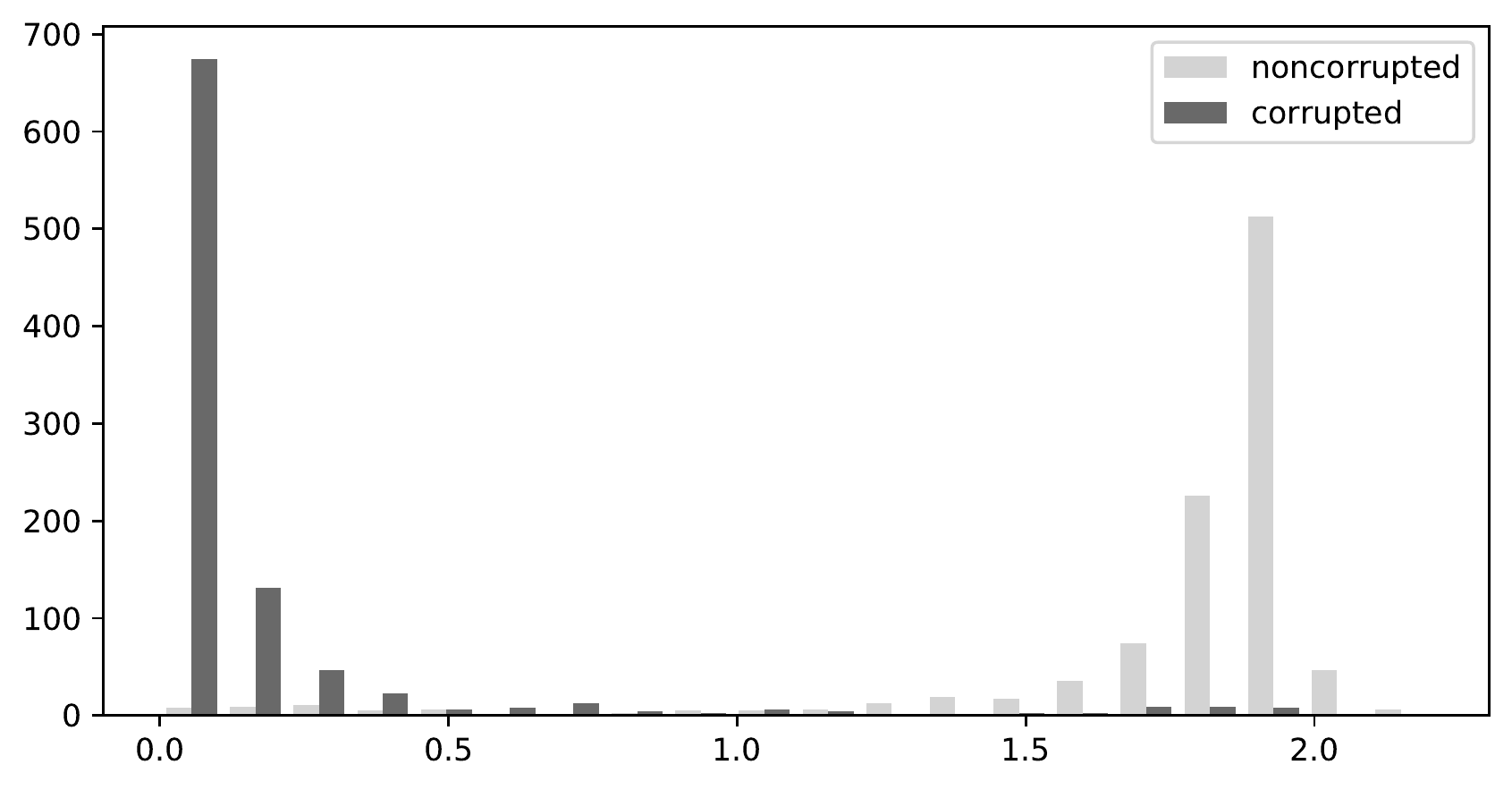}
\caption{Histogram of the $\beta$ values on AD for $p=0.25$ and $q=0$ for NCAR for the corrupted and noncorrupted examples.}
\label{beta-distrib}
\end{figure}
\begin{figure}[!h]
\centering
\includegraphics[width=\linewidth]{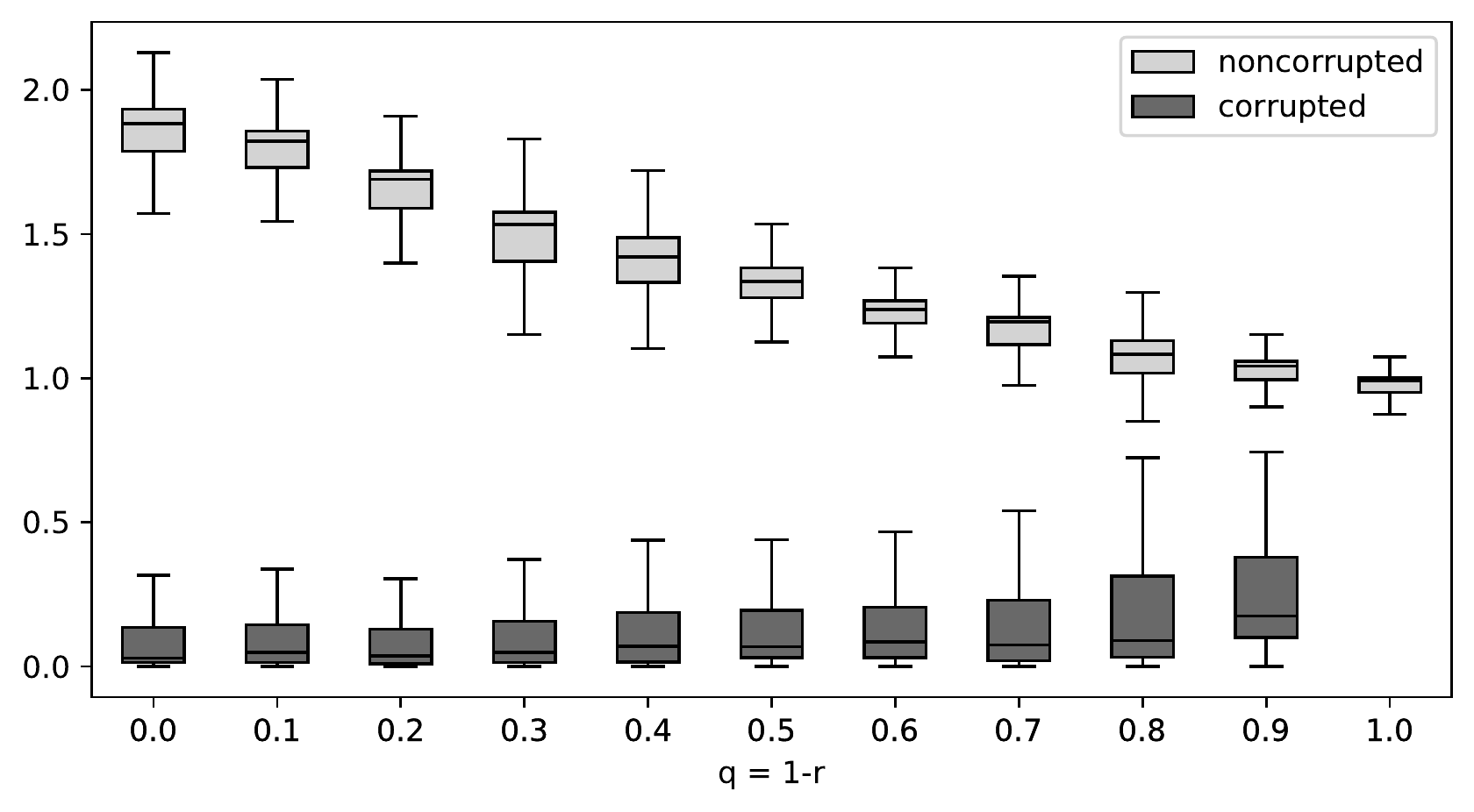}
\caption{Boxplot the $\beta$ values on AD for $p=0.25$ versus the quality, from $q=0$ to $q=1$ for NCAR.}
\label{beta-box-plot}
\end{figure}

Figure \ref{beta-distrib} shows the histogram of the weights assigned to each untrusted example either corrupted or not. It is clear that the proposed method is able to detect corrupted and noncorrupted labels from the untrusted dataset. Figure \ref{beta-box-plot} confirms this behavior when varying the value of the quality. 
For a perfect quality, the distribution of the $\beta$ is unimodal with a median equal to one and a very narrow inter quantile range, whereas, when the quality drops, the distribution  of the $\beta$ for the corrupted labels decreases to zero.

\begin{figure}[!h]
\centering
\includegraphics[width=\linewidth]{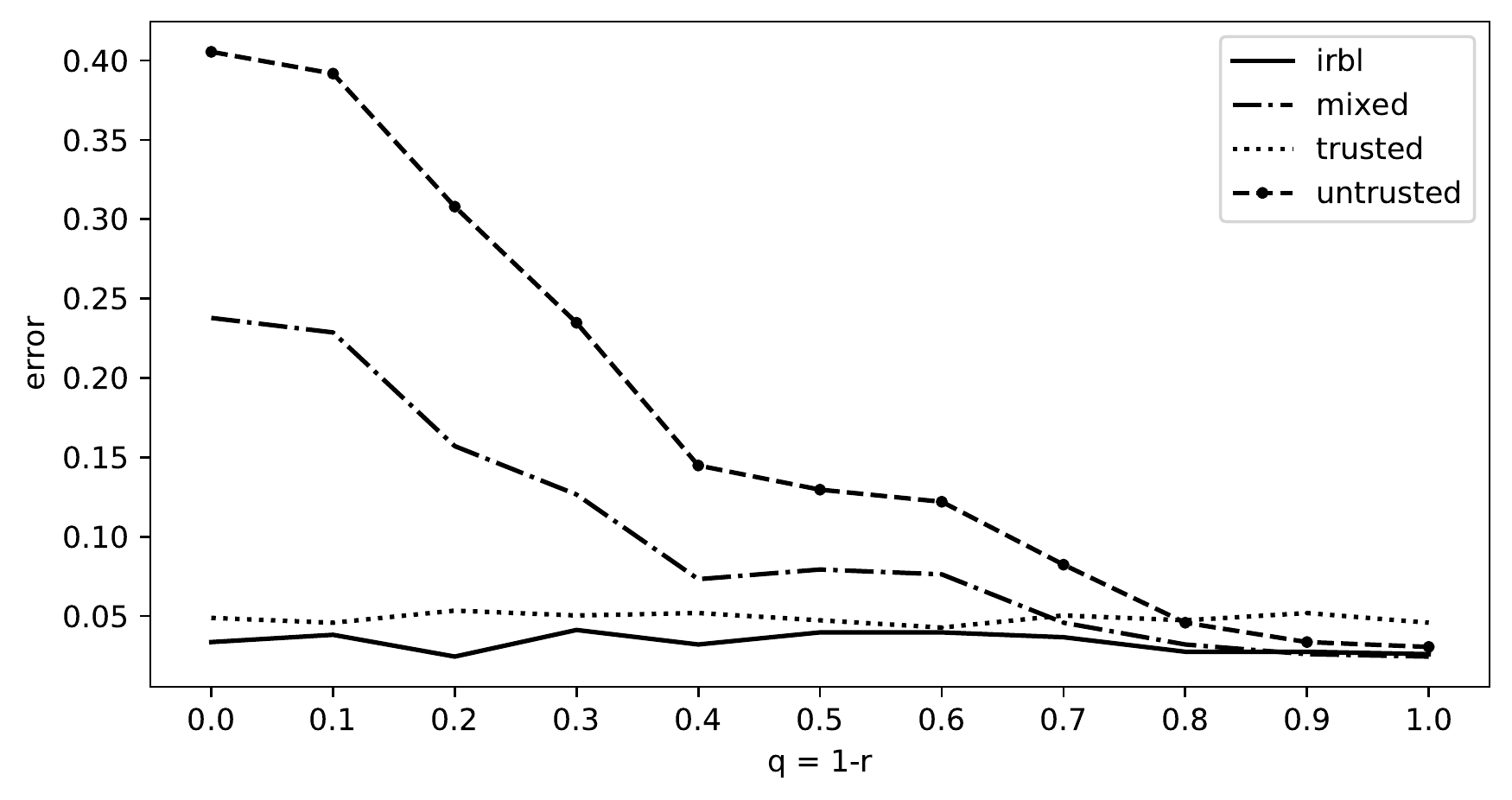}
\caption{Classification error on test set for IRBL against baselines on a full range of quality level (AD dataset, $p=0.25$, NCAR).}
\label{error-curve-simple}
\end{figure}

It is equally interesting to look at the classification error when $q$, the quality of the untrusted data, varies. Figure \ref{error-curve-simple} reports the performance for the proposed method and for the baseline competitors.
It is remarkable that the performance of our algorithm, IRBL, remains stable when $q$ decreases while the performance of the \textit{mixed} and \textit{untrusted} algorithms worsens. In addition, IRBL always obtains better performances than the trusted baseline.

\subsection{Comparison with  competitors}

\begin{figure}[!h]
\centering
\includegraphics[width=\linewidth]{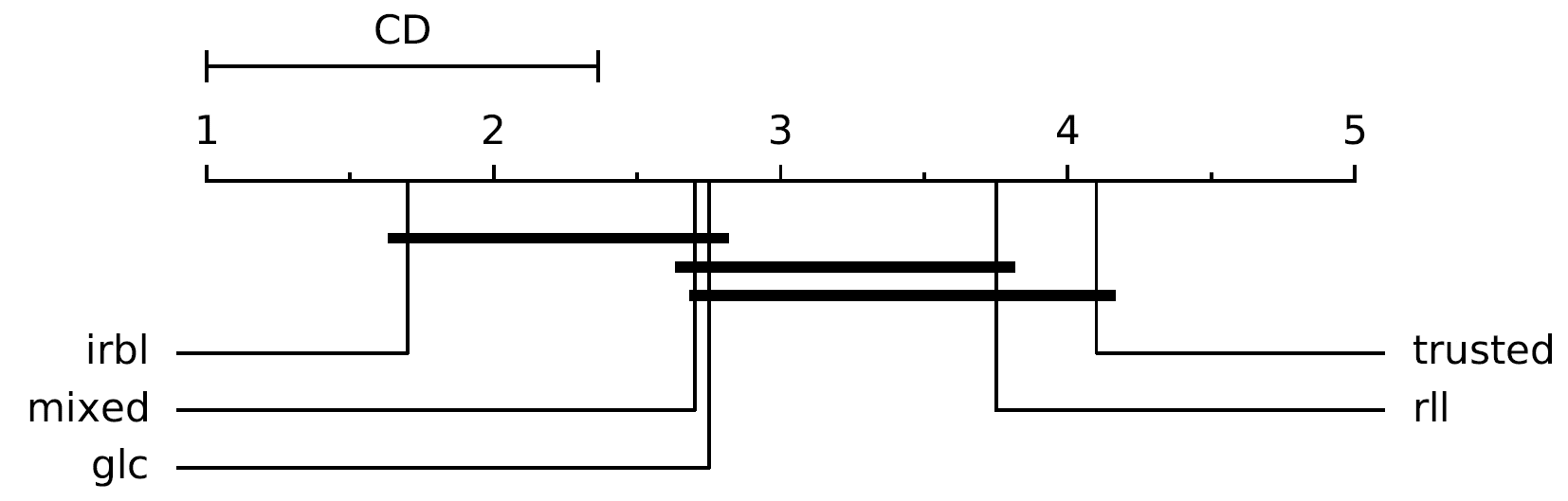}
\caption{Nemenyi test for the 20 datasets $\forall p,q$ for NCAR.}
\label{cd1}
\end{figure}

\begin{figure}[!h]
\centering
\includegraphics[width=1.0\linewidth]{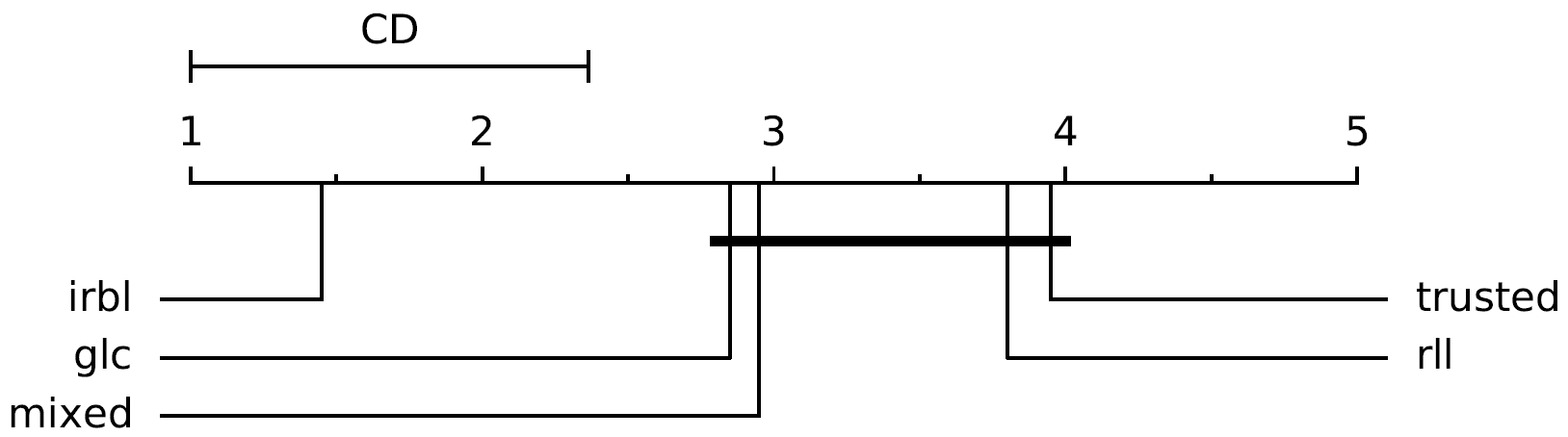}
\caption{Nemenyi test for the 20 datasets $\forall p,q$ for NNAR.}
\label{cd2}
\end{figure}

For a first global comparison, two critical diagrams are presented in Figures \ref{cd1} and \ref{cd2} which rank the various methods for the NCAR and NNAR label noise. The Nemenyi test \cite{Nemenyi62} is used to rank the approaches in terms of mean accuracy, calculated for all values of $p$ and $q$ and over all the 20 data sets described in section \ref{data_and_clf}. The Nemenyi test consists of two successive steps. First, the Friedman test is applied to the mean accuracy of competing approaches to determine whether their overall performance is similar. Second, if not, the post-hoc test is applied to determine groups of approaches whose overall performance is significantly different from that of the other groups.

\begin{table*}[!t]
\caption{Mean Accuracy (rescaled  score to be from 0 to 100) and standard deviation computed on the 20 datasets $\forall q$ for (1) NCAR and (2) NNAR. The mean ACC when using all the training data without noise is 88.65.}
\begin{center}
\begin{tabular}{|c|c|c|c|c|c|c|c|c}
\hline
 & p & trusted & irbl &  mixed   &  glc  & rll  \\ \hline
\multirow{4}{*}{(1)} & 0.02 & 72.48 $\pm$ 5.70 & {\bf 83.46 $\pm$ 3.56} & 83.40 $\pm$ 8.30 & 78.34 $\pm$ 7.94 & 77.94 $\pm$ 6.37  \\
&0.05 & 78.50 $\pm$ 4.33 & {\bf 84.94 $\pm$ 2.24} & 83.85 $\pm$ 7.35 & 81.19 $\pm$ 5.15 & 77.97 $\pm$ 6.44  \\
&0.10 & 81.40 $\pm$ 3.33 & {\bf 86.56 $\pm$ 1.68} & 85.44 $\pm$ 5.34 & 83.00 $\pm$ 3.90 & 78.98 $\pm$ 5.26  \\
&0.25 & 85.61 $\pm$ 2.39 & {\bf 87.96 $\pm$ 1.18} & 86.99 $\pm$ 2.80 & 86.27 $\pm$ 2.03 & 79.86 $\pm$ 2.61  \\ \hline
\multirow{4}{*}{(2)} &	0.02 & 72.48 $\pm$ 5.70	& {\bf 82.93	$\pm$ 3.18} & 81.30 $\pm$ 10.05 & 77.55 $\pm$ 7.78 & 75.47 $\pm$ 9.47  \\
&0.05 & 78.50 $\pm$ 4.33	& {\bf 85.34	$\pm$ 2.55} & 82.52 $\pm$ 7.72 & 80.77	$\pm$ 5.04 & 76.94 $\pm$ 6.64  \\
&0.10 & 81.40 $\pm$ 3.33	& {\bf 86.82	$\pm$ 1.45} & 84.44 $\pm$ 5.14 & 83.22	$\pm$ 4.10 & 77.95 $\pm$ 4.51  \\
&0.25 & 85.61 $\pm$ 2.39	& {\bf 88.21	$\pm$ 1.05} & 86.74 $\pm$ 2.56 & 86.56	$\pm$ 2.00 & 79.67 $\pm$ 2.70  \\ \hline
\multicolumn{2}{|c|}{Mean}  & 79.50 $\pm$ 3.94 	& {\bf 85.71 $\pm$ \underline{2.11}}	& 84.33 $\pm$ \underline{6.16}	& 82.11 $\pm$ 4.74 &	78.10 $\pm$ 5.50\\ \hline
\end{tabular}
\end{center}
\label{acc}
\end{table*}


\begin{figure*}[!h]
\begin{tabular}{ccc}
 \includegraphics[width=0.31\textwidth]{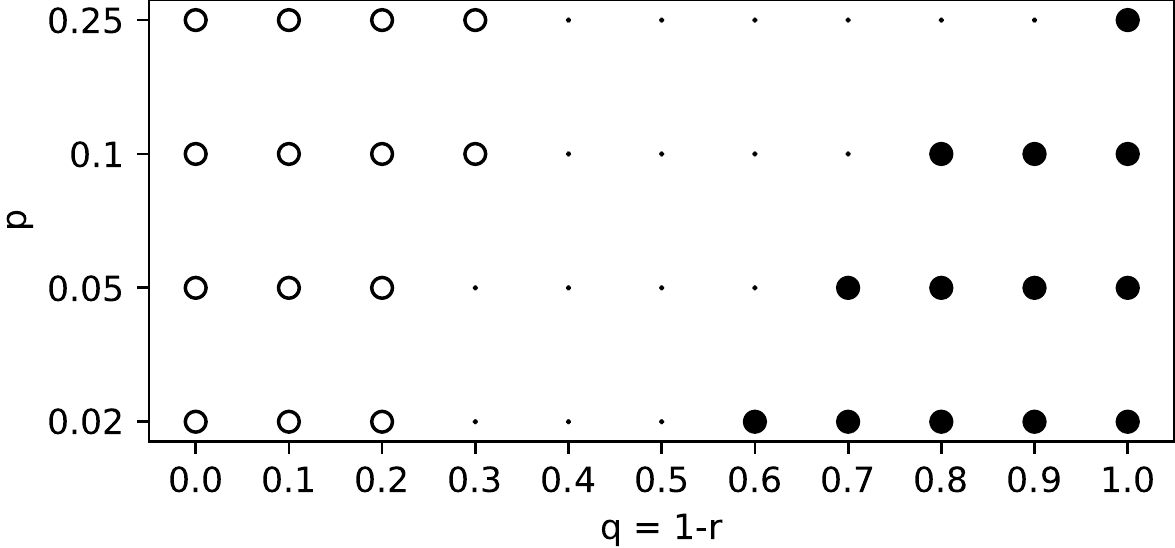} &
 \includegraphics[width=0.31\textwidth]{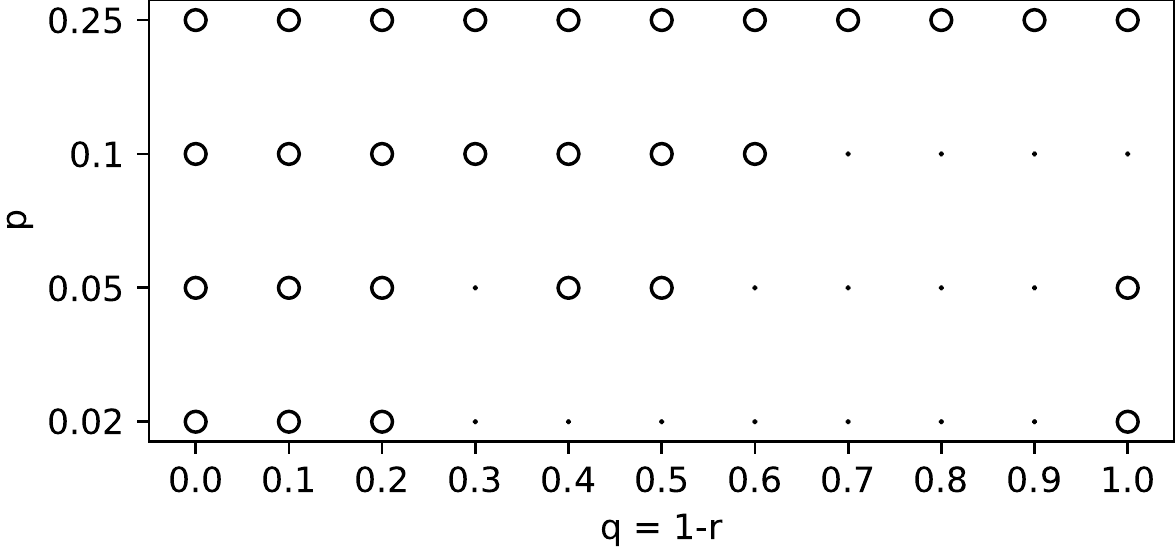} &
 \includegraphics[width=0.31\textwidth]{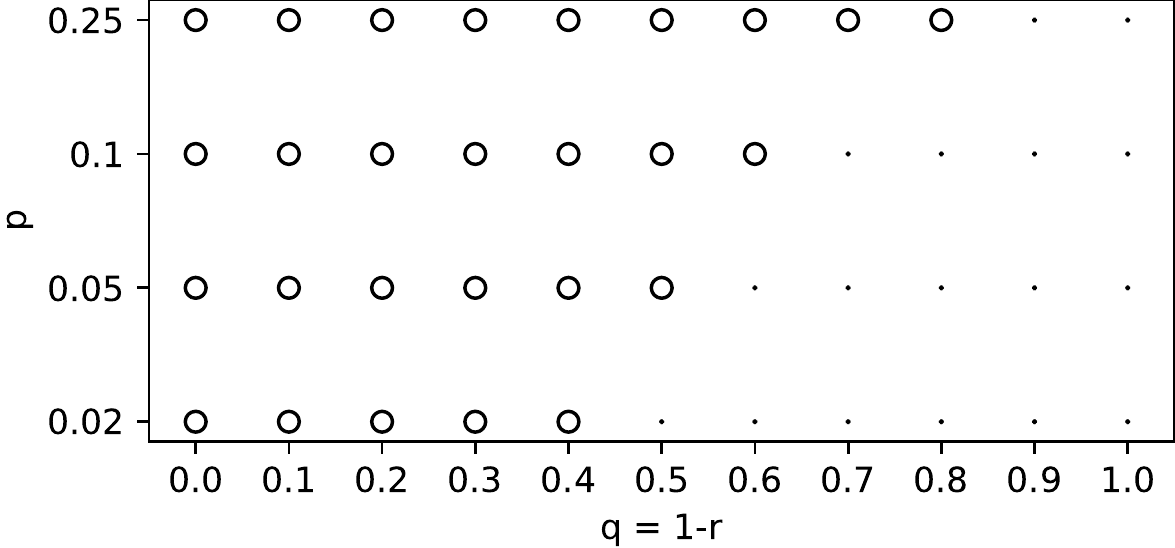} \\
 (a) IRBL vs Mixed for NCAR & (b) IRBL vs RLL for NCAR & (c) IRBL vs GLC for NCAR\\
 \includegraphics[width=0.31\textwidth]{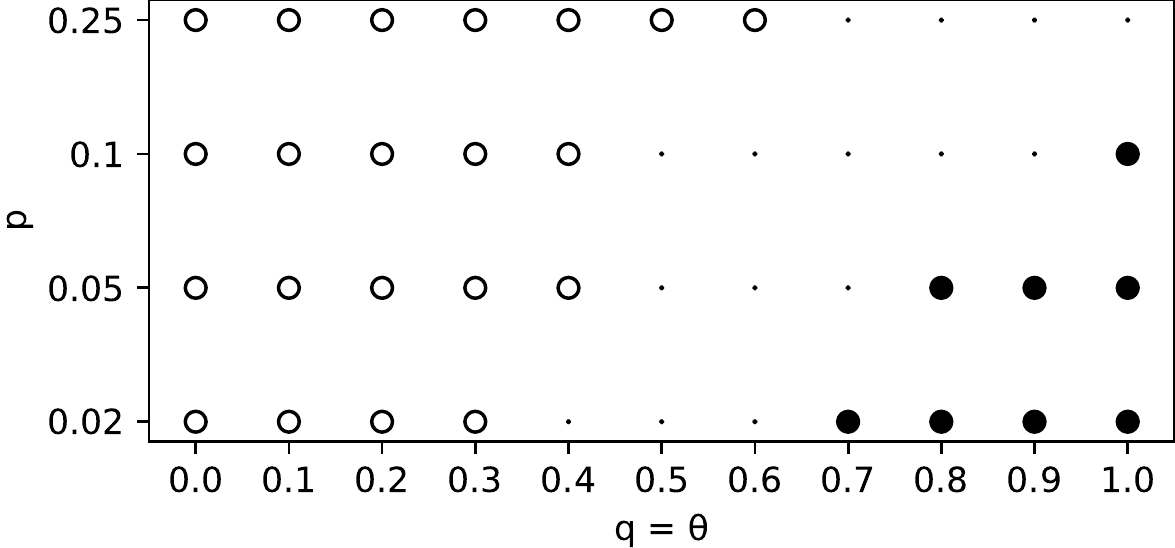} &
 \includegraphics[width=0.31\textwidth]{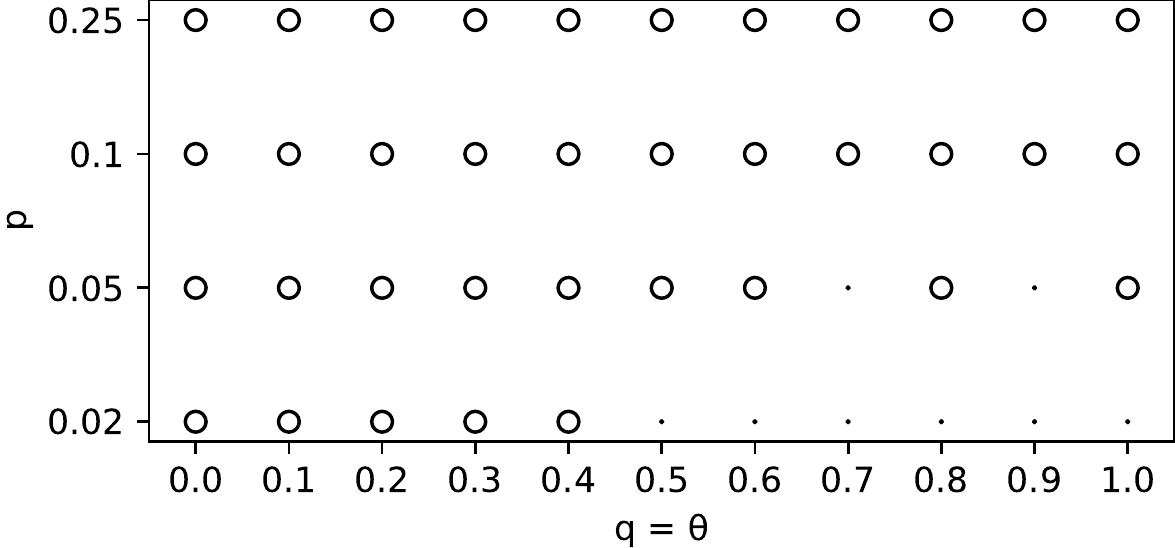} &
 \includegraphics[width=0.31\textwidth]{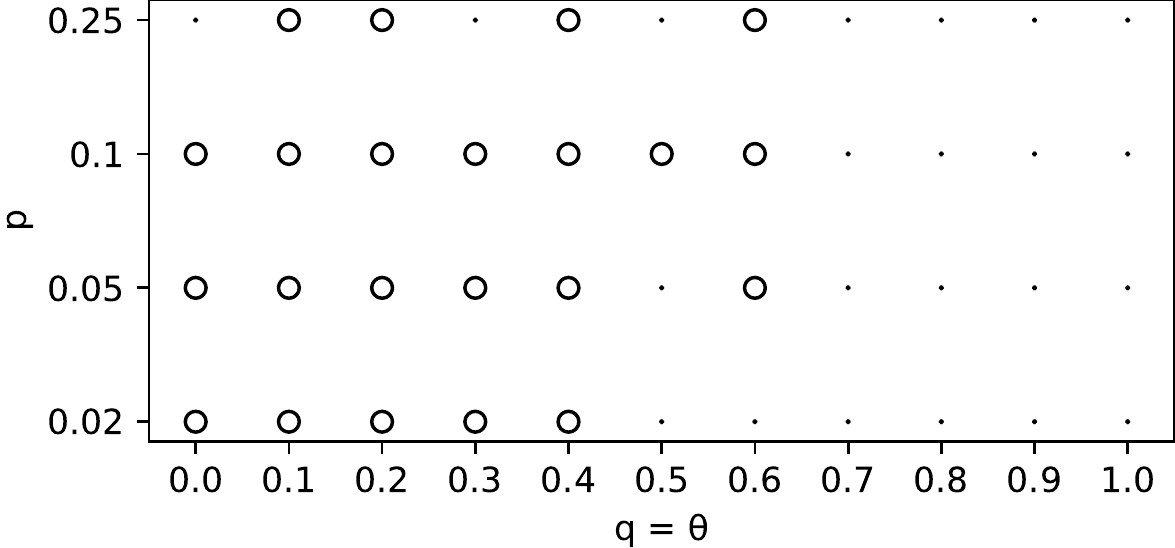} \\
 (d) IRBL vs Mixed for NNAR & (e) IRBL vs RLL for NNAR & (f) IRBL vs GLC for NNAR\\
\end{tabular}
\caption{Results of the Wilcoxon signed rank test  computed on the 20 datasets. Each figure compares IRBL versus one of the competitors. Figures a, b, c are in the case of Noisy label  Completely  at  Random  and  Figures d, e, f for the case of Noisy label Not at Random. In each figure ``$\circ$'', ``$\cdot$'' and ``$\bullet$'' indicate respectively  a win, a tie or a loss of IRBL compared to the competitors, the vertical axis is $p$ and the horizontal axis is $q$.}
\vspace{0.5cm}
\label{wilcoxon}
\end{figure*}

These figures show that the IRBL method is ranked first for the two kinds of label noise and provides better performance than the other competitors. Table \ref{acc} provides a more detailed perspective by reporting the mean accuracy and its standard deviation.
These values are computed for different values of $p$ over all qualities $q$ and  all datasets. This table also helps to see how the methods fare as compared to learning on perfect data. Overall, IRBL obtains the best results and with a lower variability.

To get more refined results, the Wilcoxon signed-rank test \cite{10.2307/3001968} is used \footnote{Here the test is used with a confidence level at 5~\%.}. It enables us to find out under which conditions -- i.e. by varying the values of p and q -- IRBL performs better or worse than the competitors.

Figure \ref{wilcoxon} presents six graphics, each reporting the Wilcoxon test that evaluates our approach against a competitor, based on the mean accuracy over the 20 datasets. The two types of label noise (see Section \ref{SupervisionDeficiencies}) correspond to the rows in Figure \ref{wilcoxon} and a wide range of $q$ and $p$ values are considered. 

Thanks to these graphs we can compare in more details our method (IRBL) with the mixed methods, as well as with RLL and GLC. Regarding the mixed method, Figures \ref{wilcoxon}  (a) and (b) return the results obtained versus varying values for $p$ and $q$.
For low quality values $q$, whatever is the value of $p$, IRBL is significantly better.
For middle values of the quality there is no winner and for high quality values and low values of $p$, the mixed method is significantly better (this result seems to be observed in \cite{Hendrycks2018} as well).
This is not surprising since at high quality values, the mixed baseline is equivalent to learning with perfect labels.

These detailed results help us to understand why, in the critical diagram in Figure \ref{cd1}, although IRBL has a better ranking, it is not significantly better than the mixed method: mainly because of the presence of high quality value cases.

Regarding the competitors RLL and GLC, Figures \ref{wilcoxon}(b), \ref{wilcoxon}(c), \ref{wilcoxon}(e)  and \ref{wilcoxon}(f) show that IRBL has always better or indistinguishable performances. Indeed, IRBL performs well regardless of the type of noise. This is an important result since it shows that we are able to deal not only with NCAR noise but also with instance dependent label noise (NNAR) which is more difficult. The  method RLL gets more ties with IRBL on NCAR than on NNAR as expected. It is noteworthy that GLC has ties with IRBL when the quality is high whatever the label noise.

To sum up, the proposed method has been tested on a large range of types and strengths of label corruptions. In all cases, IRBL has obtained top or competitive results. Consequently, IRBL appears to be a method of choice for applications where biquality learning is needed. Moreover, IRBL has no user parameter and a low computational complexity. 

\section{Conclusion}
\label{sec_conclusion}

This paper has presented an original view of Weakly Supervised Learning and has described a generic approach capable of dealing with any kind of label noise. 
A formal framework for \textit{biquality learning} has been developed where the empirical risk is minimized on the small set of trusted examples in addition to some appropriately chosen criterion using the untrusted examples. 
We identified three different ways to design a mapping function leading to three different such criteria within the \textit{biquality learning} framework. We implemented one of them: a new Importance Reweighting approach for Biquality Learning (IRBL).
Extensive experiments have shown that IRBL significantly outperforms state-of-the-art approaches, by simulating completely-at-random and not-at-random label noise over a wide range of quality and ratio values of untrusted data.

Future works will be done to extend experiments with multiclass classification datasets and other classifiers such as Gradient Boosted Trees \cite{friedman2001greedy}. An adaptation of IRBL to Deep Learning tasks with an online algorithm will be studied too.

\bibliographystyle{IEEEtran}
\bibliography{references.bib}

\end{document}